%% file: main.tex
\documentclass[10pt,conference,a4paper]{IEEEtran}
\usepackage{amsmath}
\usepackage{amssymb}
\usepackage[ruled, linesnumbered, vlined]{algorithm2e}
\usepackage{graphicx}
\usepackage{multirow}
\usepackage{subfigure}
\usepackage{multicol}
\usepackage{threeparttable}
\usepackage{booktabs}
\usepackage{color}

\newtheorem{Traffic forecasting}{Definition}[section]
\newcommand{\G}{\textsl{GLT-GCRNN}\xspace}

\newcommand{\vct}[1]{\boldsymbol{#1}} % vector
\newcommand{\mat}[1]{\boldsymbol{#1}} % matrix
\newcommand{\set}[1]{\mathcal{#1}}  % set

\ifCLASSINFOpdf
\else
\fi
\begin{document}
\title{Constructing Geographic and Long-term Temporal Graph for Traffic Forecasting}

\author{\IEEEauthorblockN{Yiwen Sun\IEEEauthorrefmark{1},
Yulu Wang\IEEEauthorrefmark{1},
Kun Fu\IEEEauthorrefmark{2},
Zheng Wang\IEEEauthorrefmark{2},
Changshui Zhang\IEEEauthorrefmark{1} and
Jieping Ye\IEEEauthorrefmark{2}}
\IEEEauthorblockA{\IEEEauthorrefmark{1}Department of Automation, Tsinghua University, \\
State Key Lab of Intelligent Technologies
and Systems, \\
Institute for Artificial Intelligence, Tsinghua University (THUAI), \\ 
Beijing National Research Center for Information Science and Technology (BNRist), Beijing, China\\
Email: \{syw17, wangyulu18\}@mails.tsinghua.edu.cn, zcs@mail.tsinghua.edu.cn}
\IEEEauthorblockA{\IEEEauthorrefmark{2}DiDi AI Labs, Beijing, China\\
Email: \{fukunkunfu, wangzhengzwang, yejieping\}@didiglobal.com}}

% \author{Yiwen Sun$^{1}$, Kun Fu$^{2}$, Zheng Wang$^{2}$, Donghua Zhou$^{3,1}$, Jieping Ye$^{2}$ and Changshui Zhang$^{4,1}$% <-this % stops a space
% \thanks{*This work was supported in part by ...}% <-this % stops a space
% \thanks{$^{1}$The Department of Automation, Tsinghua University, Beijing, 100084, China}
% \thanks{$^{2}$DiDi AI Tech, Beijing, 100084, China}
% \thanks{$^{3}$The College of Electrical Engineering and Automation, Shandong University of Science and Technology, Qingdao, Shandong, 266590, China}
% \thanks{$^{4}$The Institute for Artificial Intelligence, Tsinghua University (THUAI), the State Key Lab of Intelligent Technologies
% and Systems, the Beijing National Research Center for Information Science and Technology (BNRist), Beijing, 100084, China}
% \thanks{**Changshui Zhang is the corresponding author of this paper. {\tt\small E-mail: zcs@mail.tsinghua.edu.cn}}


\maketitle

\begin{abstract}
Traffic forecasting influences various intelligent transportation system (ITS) services and is of great significance for user experience as well as urban traffic control. 
It is challenging due to the fact that the road network contains complex and time-varying spatial-temporal dependencies.
Recently, deep learning based methods have achieved promising results by adopting graph convolutional network (GCN) to extract the
spatial correlations and recurrent neural network (RNN) to capture the temporal dependencies. 
However, the existing methods often construct the graph only based on road network connectivity, which limits the interaction between roads.
In this work, we propose Geographic and Long-term Temporal Graph Convolutional Recurrent Neural Network (\G), a novel framework for traffic forecasting that learns the rich interactions between roads sharing similar geographic or long-term temporal patterns. 
Extensive experiments on a real-world traffic state
dataset validate the effectiveness of our method by showing that \G outperforms the state-of-the-art methods in terms of different metrics.
\end{abstract}


% \fi
\IEEEpeerreviewmaketitle

\input{sections/introduction.tex}

\input{sections/related.tex}
\input{sections/method.tex}
\input{sections/experiment.tex}

\section{Conclusion and Future Work}
\label{sec:CONCLUSION}
In this work, we propose a novel deep learning based traffic forecasting framework: \G which better utilizes the GCN and LSTM for mining the spatial-temporal information simultaneously.
This paper is firstly devoted to a more comprehensive geographic and long-term temporal (GLT) graph construction method.
Furthermore, we complete the corresponding improvement of the spatial graph convolution operation and modified LSTM based on the GLT graph.
Experimental results on the large real world dataset demonstrate that \G outperforms other state-of-the art methods on the prediction performance.
Future efforts will be made to explore whether we could improve \G with the heterogeneous graph neural network and adopt our method for other tasks in spatial-temporal structured sequence prediction.


% \section*{Acknowledgment}
% This work has been supported by the...



%\bibliographystyle{IEEEtran}


% \end{thebibliography}

\bibliographystyle{IEEEtran}
\bibliography{references} 

\end{document}

%% file: sections/introduction.tex
\section{Introduction}
\label{sec:Introduction}
As one significant category of intelligent systems, ITS~\cite{dimitrakopoulos2010intelligent,figueiredo2001towards} has developed rapidly due to artificial intelligence (AI) for transportation recently. 
Putting AI to analyze the dynamic traffic patterns under the massive spatial-temporal data has obtained the promising performance in many basic and essential tasks of ITS~\cite{zhang2011data,wang2018learning,guo2019attention}.

Traffic forecasting is considered as predicting future traffic states of links (road segments) given sequential historical traffic states and the road network~\cite{cui2019traffic}. It is one of the most crucial, indispensable and challenging tasks in ITS~\cite{cui2019traffic,yu20193d,guo2019attention}. Traffic forecasting is of great importance not only for governments to monitor and control urban traffic congestion~\cite{yu2017spatio} but also for road users to plan the trip using electronic map and ride-hailing mobile apps, such as Google Map and DiDi.
Furthermore, traffic forecasting provides significant road condition information for various other important tasks in ITS, such as estimated time of arrival (ETA)~\cite{wang2018learning} and route planning~\cite{kanoulas2006finding}.
Traffic forecasting, especially the speed prediction is a particularly challenging spatial-temporal prediction problem due to complicated and dynamic traffic patterns~\cite{cui2019traffic}.

Traffic forecasting has attracted a lot of attention in the past. Existing methods can be divided into the following two categories. The first category is the classical statistical methods, such as autoregressive integrated moving average (ARIMA)~\cite{hamed1995short} and its variants.
The main disadvantage of this category of methods is that it can hardly handle high dimensional and nonlinear spatial-temporal data. 

The second category is the data-driven method analysing the complex and non-linear traffic patterns.
In this flexible category methods, traditional machine learning methods~\cite{hong2011traffic,evgeniou2000regularization,westgate2013travel,anacleto2013multivariate,davis1991nonparametric,chang2012dynamic}, outperforms the first category. Nonetheless, traditional machine learning methods are not the first choice given massive and complex spatial-temporal data though having solid mathematical foundations.

\begin{figure}[htb]
    \centering
\includegraphics[width=0.95\linewidth]{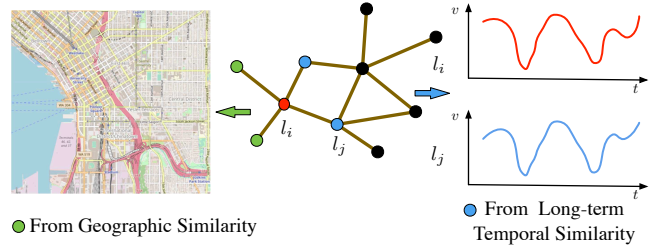}
\caption{The conceptual demonstration of the construction process of Geographic and Long-term Temporal (GLT) Graph. On the traffic graph for GCN, the neighboring nodes of each node representing a link are selected considering the geographic as well as long-term temporal similarity.}
\label{fig:sketch}
% \vspace{-5pt}
\end{figure}
Recently, deep learning~\cite{lecun2015deep} based methods~\cite{yu2017spatio,yu20193d,guo2019attention,cui2019traffic} are adept in capturing spatial-temporal dependences with large-scale datasets and become state-of-the-art. 
Compared with the relatively early dense networks for traffic forecasting~\cite{jia2016traffic,huang2014deep,lv2014traffic,chen2016learning}, the approach ~\cite{yao2018modeling} combining convolutional neural network (CNN) and RNN to model spatial-temporal correlations concurrently achieves better prediction performance.
Nevertheless, CNN are good at mining spatial relationships of two-dimensional matrices in the Euclidean space~\cite{cui2019traffic} while the road network is graph-structured
data which is not in the Euclidean space. The middle part of Fig.~\ref{fig:sketch} show a sketch map of road network topological structure.

Recent works introducing graph convolutional network (GCN) to learn the road network spatial relationships and adopting RNN or one-dimensional (1D) convolution along the time axis~\cite{cui2019traffic,yu2017spatio} become state-of-the-art. 
The graph learned by GCN of most works is constructed entirely based on the geographic information -- distance or connectivity of links. Note that distant links may also share the similar temporal patterns, for example, there is a high probability of similar congestion on two distant links near office buildings at rush hours~\cite{yu20193d}.  We think that when the state of one link is updated using the spatial GCN~\cite{niepert2016learning,kipf2016semi,zhou2017graph}, its similar links in both geographic and temporal aspects should be considered. If two links' speed distribution are resemble for a long time, this relationship should not be neglected in the traffic graph. To construct what kind of graph for traffic forecasting is a fundamental problem.

Therefore, we propose a novel geographic and long-term temporal (GLT) graph construction method. 
As illustrated in Fig.~\ref{fig:sketch}, the more comprehensive traffic graph is constructed according to both geographic and long-term temporal similarity of links.
On the traffic graph for GCN, the neighboring nodes of each link are selected considering the geographic and long-term temporal similarity.
The geographic similarity focuses on the spatial distance and the road connectivity of any two links, while the long-term temporal similarity compares the links' average speed distribution across the training dataset. 
In the light of the novel road network graph, we present the geographic and long-term graph convolution operation followed by modified RNN mining the temporal correlation.
In such a manner, our \G can adopt GCN to more effectively capture the spatial correlations of one link and its similar ones in geographical as well as long-term temporal aspects, leading to more precision traffic forecasting.

The main contributions in this paper are summarized as follows:

\begin{itemize}
    
    \item To our best knowledge, \G is the first deep learing framework which considers both the geographic and long-term temporal information when adopting GCN to mining the spatial dependency.

    \item We propose a novel road network graph construction method making use of the similarity among links in two aspects. On this graph, the neighboring nodes of each link contains the geographic as well as long-term temporal similar ones. Sufficient experiments indicate that the relationship between two distant links sharing quite similar pattens in long-term temporal is beneficial for GCN based spatial information mining.

    \item We evaluate our method on the real-world network-scale dataset which is publicly accessible~\cite{cui2019traffic} over the entirety of one year in the Greater Seattle Area. The abundant experiments demonstrate that our \G's prediction performance achieves clear improvements over other state-of-the-art methods.
    
\end{itemize}

We organize the rest of this paper as follows. Section~\ref{sec:RELATED WORK} summarizes the related works. Section~\ref{sec:METHODOLOGY} introduces the detailed description of each component as well as the overall structure of our \G. In Section~\ref{sec:EXPERIMENT}, experimental result comparisons on the real-world open dataset and traffic prediction result visualization
are presented to show the superiority of \G. Finally, we conclude this paper and discuss the future work in Section~\ref{sec:CONCLUSION}.

%% file: sections/related.tex
\section{RELATED WORK}
\label{sec:RELATED WORK}
\textbf{Traffic forecasting.} 
Traffic forecasting is one of the most essential and challenging tasks in ITS and has attracted a lot of attention in the literature. Existing methods can be summarized into the following two categories. The first category is the classical statistical methods, such as ARIMA~\cite{hamed1995short}, KARIMA~\cite{van1996combining}, SARIMA~\cite{williams2003modeling} and vector autoregression (VAR)~\cite{zivot2006vector}.
These approaches are not satisfactory because the real-world traffic data can haldly satisfy the assumptions of these methods.

The second category is the data-driven method.
Among the methods that fall into this category, traditional machine learning methods, such as Support Vector Regression (SVR)~\cite{hong2011traffic,evgeniou2000regularization}, Bayesian approaches~\cite{westgate2013travel,anacleto2013multivariate} and k-nearest neighbor (KNN)~\cite{davis1991nonparametric,chang2012dynamic}, outperforms the first category. Nonetheless, traditional machine learning methods rely heavily on handcraft feature engineering and are not skilled in mining massive and complex data.

Recently, due to the advance of deep learning in various domains~\cite{lecun2015deep,krizhevsky2012imagenet,larochelle2009exploring}, deep learning based methods~\cite{yu2017spatio,yu20193d,guo2019attention,cui2019traffic} are adept in capturing spatial-temporal dependences with large-scale datasets and become state-of-the-art. 
Among these methods, relatively early dense networks for traffic forecasting contains deep belief network (DBN)~\cite{jia2016traffic} and stacked autoencoder~\cite{lv2014traffic,chen2016learning}. \cite{huang2014deep} adopt the DBN at the bottom and a multitask (MTL) ~\cite{caruana1997multitask} regression layer at the top to take use of the sharing weights.
Compared with the traditional machine learning methods, these methods which only analyse a single region each time~\cite{guo2019deep} improve the accuracy of prediction, but the improvement is limited due to the lack of effective mining for spatiotemporal information.
Some works~\cite{zhang2017deep,ma2017learning,yu2017spatiotemporal} adopt CNN to capture the adjacent spatial correlation inspired by the rapid development of computer vision (CV) research~\cite{lecun2015deep}.
Some methods~\cite{ma2015long,yu2017deep,kong2019big} mine the temporal dependencies mainly using RNN or its varieties which are famous for sequential prediction tasks. The varieties of RNN includes long short-term memory (LSTM)~\cite{hochreiter1997long} and gated recurrent unit (GRU)~\cite{cho2014learning} networks.
As traffic forecasting is a spatial-temporal data mining problem, the approach ~\cite{yao2018modeling} combining CNN and RNN to model spatial-temporal correlations simultaneously get more accurate results for traffic forecasting.
However, CNN are good at mining spatial relationships of data with the grid structure~\cite{cui2019traffic,yu20193d} while the road network is the more complex graph-structured
data. 

\textbf{Graph Convolutional Network.} 
Graph convolution concentrates on generalizing convolution to work on structured graphs and analyze its local patterns. The graph convolution methods can be divided into two categories: the spectral methods and the spatial methods. The former methods~\cite{bruna2013spectral,henaff2015deep} are based on the spectral graph theory and \cite{bruna2013spectral} proposes the Graph Laplacian.
\cite{defferrard2016convolutional} adopts Chebyshev polynomial approximation for eigenvalue decomposition, resulting in reduced computational complexity. 
The latter spatial GCN~\cite{niepert2016learning,kipf2016semi,zhou2017graph} directly completes generalized convolution on a graph’s nodes with their neighboring nodes.
These methods are also known as adjacency matrix based GCN and the spectral methods can be regarded as a special case of the spatial GCN.  
The spatial GCN incorporating the adjacency matrix is more flexible to be applied with other network structures, furthermore, it has more potential to dispose of relatively large graph structure. Therefore, we also choose the spatial GCN for \G.

The traffic road network can be considered as a graph naturally, thus some researchers are inspired to introduce graph convolutional network (GCN) to learn the road network spatial relationships and adopt RNN or 1D convolution along the time axis~\cite{cui2019traffic,yu2017spatio}. These methods become state-of-the-art recently. ~\cite{guo2019attention} introduces the attention mechanism on the graph for GCN based traffic forecasting.
Nonetheless, the graph learned by GCN of most works is constructed entirely based on the geographic information, such as the distance and the connectivity of links. 
~\cite{yu20193d} uses a temporal only graph in terms of links' time series similarity by dynamic time warping algorithm.
Selecting the neighborhood of nodes is of great importance for the spatial method of GCN~\cite{guo2019attention}, it is essential to construct comprehensive and appropriate graph for traffic forecasting in the same light.
\G presenting a novel geographic and long-term temporal (GLT) graph construction method for road network is of great research significance and potential.

%% file: sections/method.tex
% !TEX root = ../main.tex
\section{METHODOLOGY}
\label{sec:METHODOLOGY}
We first give the definition of traffic forecasting along with the road network:

\begin{Traffic forecasting}
\textbf{Traffic forecasting}. The road network can be represented by an undirected graph
$\set{G} = \left( \set{V}, \set{E} \right)$, where $\set{V}$ is a set of nodes. $|\set{V}| = N$ represents $N$ links or sensor stations. $\set{E}$ is a set of edges which means the intersections and relevances of these links. In our study, the time interval is 5-minute, therefore, there are 288 time steps during one day. 
On the time step $t$, we denote the signal which represents the collected traffic states of all links as $\vct{x}_t \in \mathbb{R}^{N}$.
The aim of traffic forecasting is adopting a function $F(\cdot)$ which can be learned to predict next $H$-th traffic state of all links according to previous signals of $M$ time steps and the road network $\set{G}$,
\begin{equation}
\left[\vct{x}_{t-M+1}, \cdots, \vct{x}_{t}\right] \xrightarrow[\mathcal{G}]{F(\cdot)}\vct{x}_{t+H}.
\end{equation}
In this study, we focus on forecasting the representative state -- speed in the subsequent one time step, i.e. $H = 1$.
\end{Traffic forecasting}

We present the GLT Graph construction method in Section~\ref{subsec:Graph}, expound the GLT graph convolution operation and relevant modified LSTM in Section~\ref{subsec:GCN_LSTM} and introduce the overall framework of our method in Section~\ref{subsec:GLT-GCRNN}.

\subsection{Constructing the GLT Graph}
\label{subsec:Graph} 
How to construct the graph for road network is crutial to mining the dependencies among links for predicting the traffic condition. Furthermore, it is of great importance for spatial GCN to select the neighborhood nodes. A comprehensive graph could make GCN play its full role.
On the graph for GCN in this study, the neighborhood links of each link are selected considering the geographic and long-term temporal similarity.

In the geographic aspect, we adopt the adjacency matrix $\mat{A} \in \mathbb{R}^{N \times N}$ to represent the connectedness of links, following the previous work~\cite{cui2019traffic}.
The $k$-hop similar matrix in geographic aspect $\mat{S_{G}}$ can be computed:
\begin{equation}
{S_G}_{i, j}^{k}=\min \left({\left(A + I\right)}_{i, j}^{k}, 1\right).
\end{equation}

In the long-term temporal aspect, we propose a novel method. 
The long-term temporal difference matrix $\mat{Q}$ is constructed. Each element of $\mat{Q}$ is defined by the Euclidean distance of any two links' average speed distribution $\hat{\boldsymbol{v}}(i)$ and $\hat{\boldsymbol{v}}(j)$ across the training set after the combination every three time steps,
\begin{equation}
Q_{i j}=Q_{j i}=\|\hat{\boldsymbol{v}}(i)-\hat{\boldsymbol{v}}(j)\|_{2}.
\end{equation}
The combination which means averaging the speed of every three time steps lead to the result that the dimension of the link's one day speed distribution vector is reduced from 288 to 96. 
The similar matrix in the long-term temporal aspect $\mat{S_{LT}}$ is constructed by preserving top $\gamma$ the closest links for each link:

\begin{equation}
{S_{LT}}_{i, j}=\begin{cases}1, & Q_{i, j} \in Q_{i, :} \ \text{top} \ \gamma \ \text{small elements}\\ 
 0, & \text{otherwise}\end{cases}
\end{equation}
where $\gamma$ is a hyper-parameter to adjust the number of preserved long-term temporal similar links.

\begin{figure}[htb]
    \centering
\includegraphics[width=0.7\linewidth]{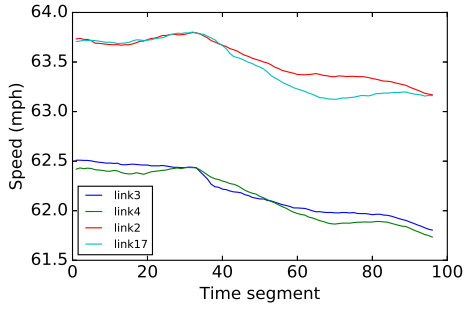}
\caption{Two representative groups of long term temporal similar links.}
\label{fig:LT_similar}
% \vspace{-5pt}
\end{figure}

The $k$-hop geographic and long-term temporal similar matrix $\mat{S_{GLT}}$ can be formulated as below:

\begin{equation}
\mat{S_{GLT}}^{k} = \mat{S_G}^{k} + \mat{S_{LT}}.
\end{equation}
In the light of $\mat{S_{GLT}}$, we could realize the effective interaction of geographic and long-term temporal similar links in the process of GCN updating features. We show two representative groups of long-term temporal similar links in Fig.~\ref{fig:LT_similar} to demonstrate the importance of $\mat{S_{LT}}$ visually. The open dataset adopted will be introduced in Section~\ref{subsec:EXPERIMENT}.
As illustrated in Fig.~\ref{fig:LT_similar}, link 3 (ID) is the most Long-term Temporal (LT) similar link of link 4 while they are also geographic similar.
Link 2 is the most LT similar link of link 17, nevertheless, they are not geographic similar which is ignored by $\mat{S_{G}}$. 

With reference to the established traffic flow theory~\cite{DaganzoThe}, we also adopt free-flow reachable matrix $\mat{S_F}$ which is the same as~\cite{cui2019traffic}.
For two elements links with ID=$i$ and ID=$j$, $V_{i, j}$ is the free-flow speed which is the traffic flow speed unaffected by upstream or downstream conditions~\cite{HeWill} or the average speed without adverse conditions, such as congestion~\cite{cui2019traffic}.
$D_{i, j}$ is the real roadway distance from link $i$ to $j$.
Free-Flow Reachable Matrix $\mat{S_F}$ is defined:

\begin{equation}
{S_F}_{i, j}=\begin{cases} 1, & V_{i, j} m \Delta t \geq D_{i, j} \\ 0, & \text {otherwise}\end{cases}
\end{equation}
where $\Delta t$ is the time quantum duration whose value is often chosen as a relatively big number, such as 20 minutes and $m$ is the considered time interval number when calculating the distance.

Then, the $k$-hop ultimate similar matrix for GCN $\mat{S_{U}}^{k}$ can be calculated:
\begin{equation}
\mat{S_{U}}^{k} = \mat{S_{GLT}}^{k} \odot \mat{S_F},
\end{equation}
where $\odot$ is the Hadamard product operator. The main role of $\mat{S_F}$ is to filter out the links that are far away even though the traffic condition is free flow.

\subsection{GLT Graph Convolution Operation and Modified LSTM}
\label{subsec:GCN_LSTM}
We then modify the graph convolution operation and LSTM according to constructed GLT graph.
The spatial GCN is adopted to extract localized features of the input $\vct{x}_{t}$ which is similar to the previous work~\cite{cui2019traffic}.
Specifically speaking, the product of a trainable weight matrix $\mat{W_g}^{k}$ Hadamard producted by $\mat{S_U}^{k}$ and $\vct{x}_{t}$ to realize the GLT graph convolution operation.
\begin{equation}
\vct{g}_{t}^{k}=\left(\mat{W_g}^{k} \odot \mat{S_U}^{k} \right) \vct{x}_{t},
\end{equation}
where $\vct{g}_{t}^{k}$ is the feature extracted by GLT graph convolution operation.
To enrich the feature space for capturing spatial correlation effectively, $\vct{g}_{t}^{k}$ extracted according to various hops $\mat{S_{U}}^{k}$ , i.e. 1 to $K$ are concatenated as:
\begin{equation}
\mat{G}_{t}^{K}=\left[\vct{g}_{t}^{1}, \vct{g}_{t}^{2}, \ldots, \vct{g}_{t}^{K}\right].
\end{equation}

The $\mat{G}_{t}^{K}$ considering various receptive field replaces the input of following LSTM for mining the dynamic temporal dependencies. Another improvement for the vanilla LSTM is as follows which is with reference to~\cite{cui2019traffic}.
Because the LSTM cell state of each link in the graph should also be affected by its neighboring links' cell states, we design a cell state gate as follows in the LSTM cell. 
\begin{equation}
C_{t-1}^{*}=\mat{W_C} \odot \mat{S_{U}}^{K} \cdot C_{t-1},
\end{equation}
where $\mat{W_C}$ is the weight matrix which is constrained by $\mat{S_{U}}^{K}$ to measure the influence degree of neighboring links' cell states. $C_{t-1}^{*}$ replaces the $C_{t-1}$ when calculating the final cell state.

\subsection{Overall Framework of \G}
\label{subsec:GLT-GCRNN}

\begin{figure}[htb]
    \centering{}{}
\includegraphics[width=0.99\linewidth]{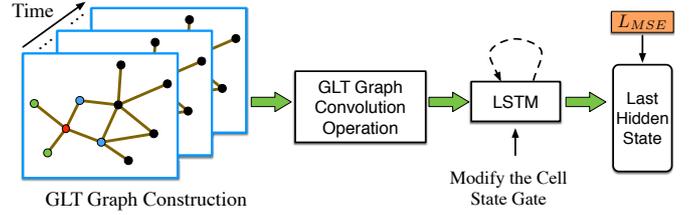}
\caption{The overall architecture of \G. Our framework consists of three components: (1) the GLT graph construction. (2) GLT graph convolution operation for mining the spatial correlation. (3) Modified LSTM to capture the time series dependency.}
\label{fig:framework}
% \vspace{-5pt}
\end{figure}

We present the \G deep learning framework for traffic forecasting, as shown in Fig.~\ref{fig:framework}.
\G consists of three components: (1) the GLT graph construction which is introduced in Section~\ref{subsec:Graph}. 
This novel graph construction method ensures that each link's neighboring nodes contrain the long-term temporal similar links besides geographic simialr ones. This road network graph of comprehensive information is beneficial for spatial GCN to aggregate neighboring information;
(2) the GLT graph convolution operation is the main module to learn the spatial correlation of the center link and its neighboring links; (3) the modified LSTM is adopted to capture the temporal dependency of the link's traffic state considering the neighboring nodes.
The components: (2) and (3) are presented in detail in Section~\ref{subsec:GCN_LSTM}.

The final time step $t$ hidden state $h_t$ of the modified LSTM is served as the predicted value for traffic forecasting. For the objective function, we choose the Mean Squared Error (MSE) function which is defined as 
\begin{equation}
L_{MSE} =  \frac{1}{n} \sum_{i = 1}^{n}{ \left( y_{i} - y_{i}' \right)^2} , 
\label{eq:MAPE}
\end{equation}
where $y_i$ is the ground-truth traffic state of one link of one sample and $y_{i}'$ is the predicted one. $n$ represents the product of the total number of samples and the number of links in the road network. The MSE function minimized by backpropagation (BP) in our study is popular for traffic prediction.

%% file: sections/experiment.tex
% !TEX root = ../main.tex
\section{EXPERIMENT}
\label{sec:EXPERIMENT}

The evaluation is on the open network-scale real-world dataset. The dataset, the competing methods, the implementation details and the experimental results will be introduced successively.

\subsection{Dataset}
\label{subsec:EXPERIMENT}
The real-world dataset which is adopted to demonstrate the advantages of our framework is publicly accessible~\cite{cui2019traffic}.
% \footnote{https://github.com/zhiyongc/Seattle-Loop-Data}
The traffic state data is collected from inductive loop detectors deployed on four connected freeways: I-5, I-405, I-90, and SR-520~\cite{cui2019traffic}.
The data statistics are summarized in Table~\ref{tbl:dataset}.
For the dataset, we use $70\%$ of the data as training set and $20\%$ and $10\%$ as validation set and test set, respectively.  
The speed limit (60mph) is adopted as the free-flow speed and the distance adjacency matrices as well as free-flow reachable matrix are calculated based on the road network's topology and characteristics, which are the same as~\cite{cui2019traffic}.
\begin{table}[htb]
\centering
    \caption{Statistics of the Real-world Traffic Forecasting Dataset}
\label{tbl:dataset}
\begin{tabular}{c c c c} 
 \toprule
  location               & time span              & \# sensor         & interval \\
 \midrule
  Greater Seattle Area   & 1 year (2015)          &  323              & 5-minute \\
\bottomrule
\end{tabular}
\end{table}

\subsection{Competing Methods}
We compare the proposed \G with the following baselines including the state-of-the-art method.

(1) ARIMA: Auto-Regressive Integrated Moving Average model~\cite{hamed1995short} which is a representative method of the classical statistical methods. 

(2) SVR~\cite{hong2011traffic,evgeniou2000regularization}: Support Vector Regression.

(3) FNN: Feed forward neural network which is also known as the multilayer perceptron (MLP) with two hidden layers.

(4) LSTM: Long Short-Term Memory recurrent neural network~\cite{hochreiter1997long}.

(5) DiffGRU:  diffusion convolutional gated recurrent network whose gate units are defined based on diffusion convolution is first proposed by~\cite{li2017diffusion} for traffic forecasting. 
Since the graph is undirected in our study, the diffusion convolution is replaced with the spectral graph convolution in DiffGRU that is the consistent with~\cite{cui2019traffic}.

(6) Conv+LSTM: a 1D convolution layer (kernel size=5 and stride=2) with two channels followed by an LSTM layer which is also a baseline used by~\cite{cui2019traffic}. 

(7) SGC+LSTM: a spectral graph convolution layer~\cite{henaff2015deep} stacked with an LSTM layer.

(8) LSGC+LSTM: stacking a one-layer localized spectral graph convolution layer~\cite{defferrard2016convolutional} which is stacked with an LSTM layer just like SGC+LSTM.

(8) TGC-LSTM: The method using spatial GCN to learn the road network spatial relationships and adopting LSTM along the time axis~\cite{cui2019traffic} becomes state-of-the-art.

For the comparation of all these methods quantificationally, we take Root Mean Square Error (RMSE), Mean Absolute Percentage Error (MAPE) and Mean Absolute Error (MAE) which are used widely in traffic prediction tasks~\cite{yu20193d} as the evaluation metrics.
Their computations are as followed:
\begin{equation}
\begin{aligned}
\text{RMSE} &= \sqrt{\frac{1}{n} \sum_{i = 1}^{n}{ \left( y_{i} - y_{i}' \right)^2}},\\
\text{MAPE} &= {\frac{\hbox{1}}{n}} \sum_{i = 1}^{n}{\frac{\left\vert y_{i} - y_{i}' \right\vert}{y_{i}}},\\
\text{MAE} &= \frac{1}{n} \sum_{i = 1}^{n}{ \left\vert y_{i} - y_{i}' \right\vert}.
\end{aligned}
\end{equation}

\subsection{Implementation Details}
\label{sec:implementation}
\G are implemented in PyTorch~\cite{paszke2019pytorch}, and the training process is accelerated on a single NVIDIA P40 GPU. We set the maximal epoch number to 200 with the early stopping mechanism.
The dimensions of the hidden states of \G are set as the amount of the nodes in the GLT graphs which is equal to the number of sensors. The size of hops $K$ in the GLT graph convolution is set as 3.
The initial learning rate is set to $10^{-5}$ and the mini-batch size is 10.
All the parameters are jointly trained using RMSProp~\cite{tieleman2012lecture} optimizer, which could solve the gradient exploding and vanishing problems. The $\alpha$ and $\epsilon$ of RMSProp is set as 0.99 and $10^{-8}$.
The settings of the above parameters are consistent with~\cite{cui2019traffic} for comparing fairly. 
For \G, the hyper-parameter $\gamma$ of GLT graph selected by the results on validation set is 3.

\subsection{Experimental Results}
We list the results of baselines, ARIMA to TGC-LSTM (the result numbers from ~\cite{cui2019traffic}) as well as our \G with the same experiment settings in Table~\ref{tbl:result}. We repeat the training process of \G for 10 times and report the mean value of the models' test results.
The best scores in terms of three metrics are marked by bold font.

\begin{table}[htb]
\centering
    \caption{Traffic Forecasting Result Comparison on the Network-scale Dataset}
\label{tbl:pickup}
\begin{tabular}{c c c c} 
 \toprule
 & RMSE (mph) & MAPE (\%)  & MAE (mph) \\
 \midrule
 ARIMA     & 10.65 & 13.85 & 6.10  \\
 SVR       & 11.12 & 14.39 & 6.85  \\
 FNN       & 7.83  & 10.19 & 4.45  \\
 LSTM      & 4.97  & 6.83  & 2.70  \\
 DiffGRU   & 8.22  & 11.18 & 4.64  \\
 Conv+LSTM & 5.02  & 6.79  & 2.71  \\
 SGC+LSTM  & 4.80  & 6.52  & 2.64  \\
 LSGC+LSTM & 6.18  & 7.51  & 3.16  \\
 TGC-LSTM  & 4.63  & 6.01  & 2.57  \\
 \G (ours) & \textbf{3.59} & \textbf{5.90} & \textbf{2.45} \\

\bottomrule
\end{tabular}
\label{tbl:result}
\end{table}

As illustrated in Table~\ref{tbl:result}, the following results can be summarized:
(1) the proposed \G outperforms all the competitors regarding to all metrics. 
\G improves the accuracy to predict the future traffic condition, even compared with state-of-the-art TGC-LSTM. 
\G reduces $22.46\%$ RMSE, reduces $1.83\%$ MAPE and reduces $4.67\%$ MAE respectively in contrast to TGC-LSTM.
This could be explained that the key GLT graph of \G is constructed considering both geographic and long-term temporal information, therefore the spatial graph
convolution operation and LSTM which are modified based on GLT graph could capture the spatial-temporal correlations more effectively;
(2) it can be observed that deep learning based methods are superior to the representative traditional statistical method and traditional machine learning method, i.e. ARIMA and SVR;
(3) there is also a performance gap between the basic FNN and LSTM as well as Conv+LSTM which are good at mining the time series imformation. GCN do good to capture the spatial correlation for the road network with graph structure by comparing the predictive capability of GCN based methods, such as SGC+LSTM, TGC-LSTM, \G, and LSTM.

\subsection{Influence of Hyper-parameter}
\begin{figure}[htb]
    \centering
\includegraphics[width=1.0\linewidth]{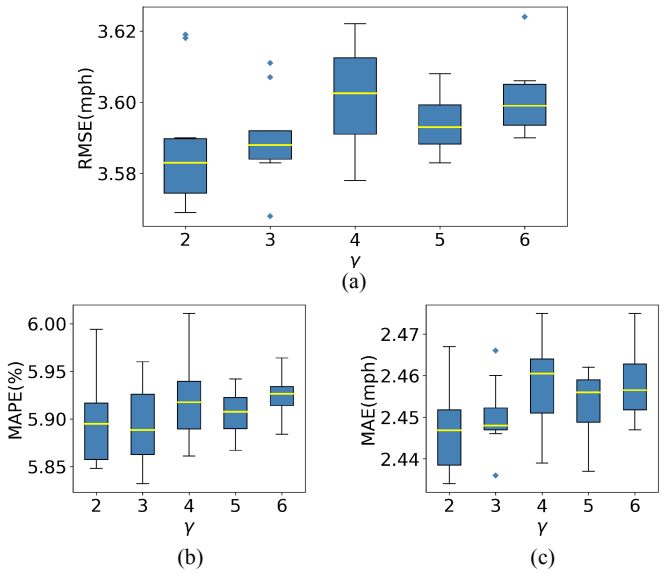}
\caption{The influence of hyper-parameter $\gamma$: (a) Regarding to RMSE. (b) Regarding to MAPE. (c) Regarding to MAE. \G generally outperforms all the competitors, which demonstrates the robustness of our framework.}
\label{fig:hyper}
% \vspace{-10pt}
\end{figure}

To study the influence of different hyper-parameter values of \G, we further train 50 models with different values of the main hyper-parameter: $\gamma$ (10 models for each $\gamma$ value). We plot the prediction error boxplots regarding to three metrics in Fig.~\ref{fig:hyper}. 
$\gamma$'s main funtion is to balance the trade-off between the geographic and the long-term temporal information in the road network graph for GLT-GCN. Greater $\gamma$ means more retention ratio of the temporal information.
We find that the RMSE, MAPE and MAE varies slightly under different $\gamma$.
The moderate $\gamma = 3$ achieves the best performance considering the Fig. \ref{fig:hyper} (a), (b) and (c) comprehensively. 

Furthermore, \G is robust enough in terms of the multiple training random error through observing three boxplots.
\G achieves better performance than state-of-the-art TGC-LSTM from $\gamma = 2$ to $6$, which demonstrates that the superiority of \G is not sensitive to the $\gamma$ hyper-parameter.

\subsection{Traffic Forecasting Visualization}
\begin{figure}[htb]
    \centering
\includegraphics[width=0.95\linewidth]{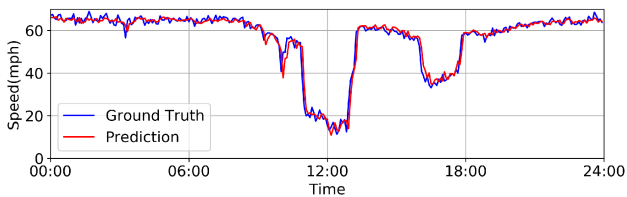}
\caption{The traffic forecasting visualization of \G. The link's ID is 190 for the visualization and the date is 2015-12-27.}
\label{fig:visual}
% \vspace{-10pt}
\end{figure}
For visualizing the prediction performance of our \G, we randomly select one link's sequential predicted traffic speed and ground truth during one day. The Fig.~\ref{fig:visual} demonstrates that \G could effectively capture the changing trend of the traffic condition at different time quantums of the day.
The prediction of sequential traffic speeds and the real ones match very accurately which implies that \G is useful in analyzing the complex traffic patterns.